\documentclass[10pt,twocolumn,letterpaper]{article}

\usepackage{cvpr}
\usepackage{times}
\usepackage{epsfig}
\usepackage{graphicx}
\usepackage{amsmath}
\usepackage{amssymb}
\usepackage{multirow}

\cvprfinalcopy 

\begin{document}

\title{Learning Single-Image Depth from Videos using Quality Assessment Networks}

\author{
Weifeng Chen\textsuperscript{1,2}\kern10pt Shengyi Qian\textsuperscript{1}\kern10pt Jia Deng\textsuperscript{2} 
\vspace{2mm}
\\
\begin{minipage}{\columnwidth}
	\centering
	\textsuperscript{1}University of Michigan, Ann Arbor
	{\tt\small \{wfchen,syqian\}@umich.edu}\\
\end{minipage}
\begin{minipage}{\columnwidth}
	\centering
	\textsuperscript{2}Princeton University\\
	{\tt\small jiadeng@cs.princeton.edu}\\
\end{minipage}  
}

\maketitle

\begin{abstract}
Depth estimation from a single image in the wild remains a 
challenging problem. One main obstacle is the lack of high-quality training
data for images in the wild. In this paper we propose a
 method to automatically 
generate such data through Structure-from-Motion (SfM) on Internet videos. The core of this method is a
Quality Assessment Network that identifies high-quality
reconstructions obtained from SfM. Using this method, we collect
single-view depth training data from a large number of YouTube videos and construct a new
dataset called \emph{YouTube3D}. Experiments show that YouTube3D is useful 
in training depth estimation networks and advances the state of the art of single-view
depth estimation in the wild. Project website: \url{https://pvl.cs.princeton.edu/youtube3d}.
\end{abstract}

\begin{figure*}[t]
    \begin{center}
        \includegraphics[width=0.9\linewidth]{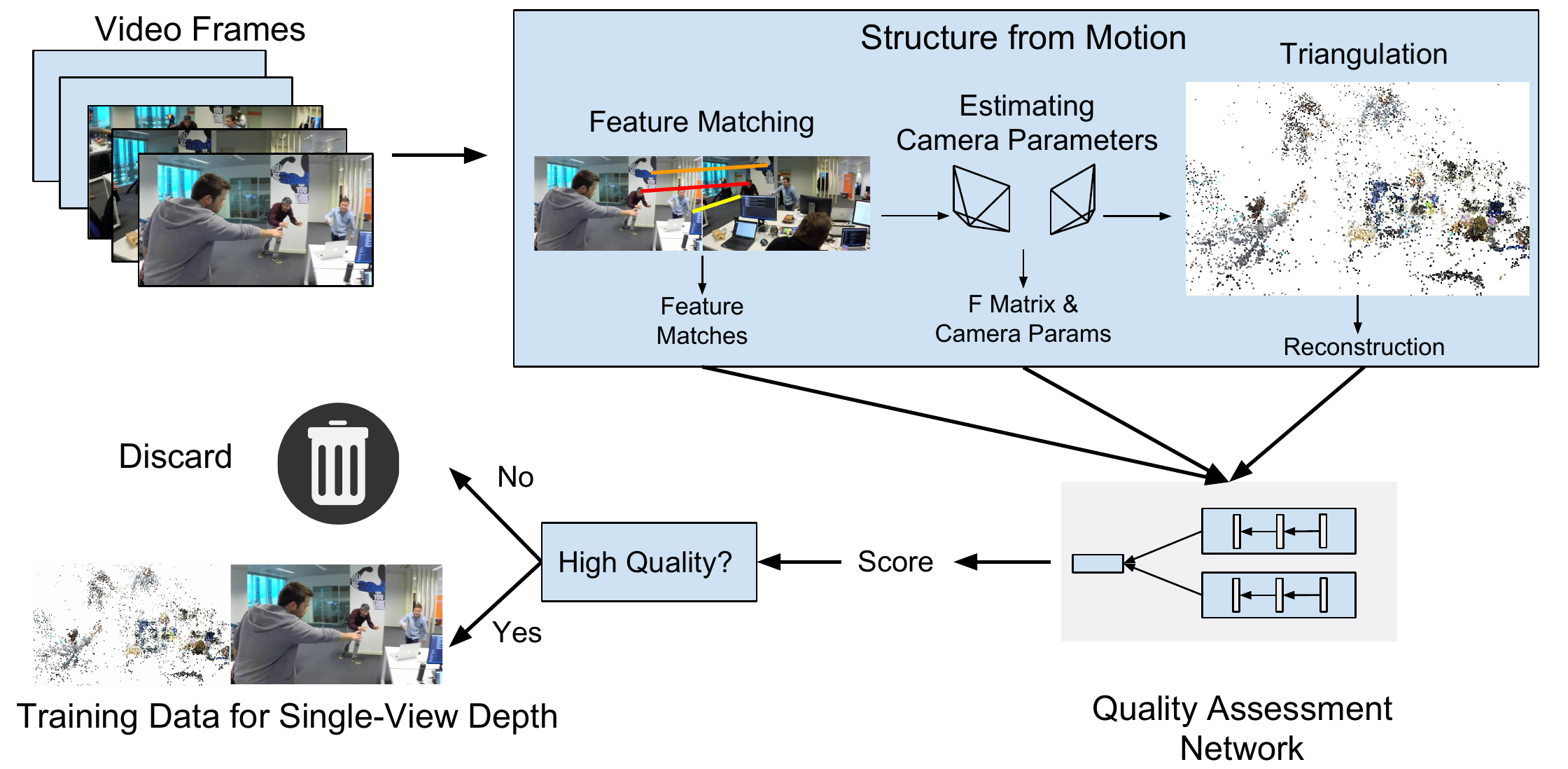}
    \end{center}
    \caption{An overview of our data collection method. Given an arbitrary video, we follow standard steps of structure-from-motion: extracting feature points and matching them across frames, estimating the camera parameters, and performing triangulation to obtain a reconstruction. A Quality Assessment Network (QANet) examines the operation of the SfM pipeline and assigns a score to the reconstruction. If the score is above a certain threshold, this reconstruction is deemed of high quality, and we use it as single-view depth training data. Otherwise, the reconstruction is discarded.}
    \label{fig:teaser}
\end{figure*}

\section{Introduction}
\label{sec:introduction}
This paper addresses the problem of single-image depth estimation, a
fundamental computer vision problem that remains challenging. Despite
significant recent
progress~\cite{wang2018learning,godard2016unsupervised,roy2016monocular,li2015depth,hane2015direction,liu2015deep,wang2015towards,xie2016deep3d,eigen2015predicting,li2016learning,kuznietsov2017semi,xu2017multi,laina2016deeper,garg2016unsupervised,ummenhofer2017demon,zhou2017unsupervised,kar2017learning,vijayanarasimhan2017sfm},
current systems still perform poorly on arbitrary images in the
wild~\cite{chen2016single}. One major obstacle is the lack of diverse
training data, as most existing RGB-D datasets were collected via
depth sensors and are limited to rooms
~\cite{silberman2012indoor,dai2017scannet,chang2017matterport3d} and
roads~\cite{geiger2013vision}. As shown by recent
work~\cite{chen2016single}, systems trained on such data are unable to
generalize to diverse scenes in the real world.

One way to address this data issue is crowdsourcing, as demonstrated
by Chen et al.~\cite{chen2016single}, who crowdsourced human
annotations of depth and constructed a dataset called
``Depth-in-the-Wild (DIW)'' that captures a broad range of scenes. One
drawback, though, is that it requires a large amount of manual
labor. Another possibility is to use synthetic
data~\cite{Butler:ECCV:2012,mayer2016large,richter2016playing,krahenbuhl2018free}, but it
remains unclear how to automatically generate scenes that match the
diversity of real-world images.

In this paper we explore a new approach that automatically collects
single-view training data on natural in-the-wild images, without the
need for crowdsourcing or computer graphics. The idea is to
reconstruct 3D points from Internet videos using Structure-from-Motion
(SfM), which matches feature points across video frames and infers
depth using multiview geometry. The reconstructed 3D points can then
be used to train single-view depth estimation. Because there is a virtually unlimited
supply of Internet videos, this approach is especially
attractive for generating a large amount of single-view training data.

However, to implement such an approach in practice, 
there remains a significant technical hurdle---despite great
successes~\cite{agarwal2011building,heinly2015reconstructing,schoenberger2016sfm,schoenberger2016mvs,mur2015orb},
existing SfM systems are still far from reliable when applied to
arbitrary Internet videos. This is because SfM operates by matching features across
video frames and reconstructing depth assuming a static scene, but
feature matches are often unreliable and scenes often contain moving
objects, both of which cause SfM to produce erroneous
3D reconstructions.  That is, if we simply apply an off-the-shelf SfM
system to arbitrary Internet videos, the resulting single-view
training data will have poor quality.

 To address this issue, we propose to train a deep network to automatically assess the quality of a SfM
 reconstruction. The network predicts a quality score of a SfM
 construction by examining the operation of the entire SfM
 pipeline---the input, the final output, along with intermediate
 outputs generated inside the pipeline. We call this network a
 \emph{Quality Assessment Network (QANet)}. Using a QANet, we filter
 out unreliable reconstructions and obtain high-quality single-view
 training data. Fig.~\ref{fig:teaser} illustrates our data collection method. 

 It is worth noting that because Internet videos are
 virtually unlimited, it is sufficient for a QANet to be able to reliably identify
 a small proportion of high-quality reconstructions. 
 In other words, high precision is necessary but high recall is
 not. This means that training a QANet will not be hopelessly
 difficult because we do not need to detect \emph{every} good
 reconstruction, only \emph{some} good reconstructions.

We experiment using Internet videos in the wild. Our experiments 
show that with QANet integrated with SfM, we can
 collect high-quality single-view training data from unlabeled
 videos, and such training data can supplement existing data to
significantly improve the performance of single-image depth estimation. 

Using our proposed method, we constructed a new dataset called
YouTube3D, which consists of 795K in-the-wild images, each associated
with depth annotations generated from SfM reconstructions filtered by a
QANet. We show that as a standalone training set for in-the-wild depth
estimation, YouTube3D is superior to existing datasets constructed with human
annotation. YouTube3D also outperforms MegaDepth~\cite{li2018megadepth}, a recent datatset automatically
collected through SfM on Internet images. In addition, we show that as
a supplement to existing
RGB-D data, YouTube3D advances the state-of-the-art of single-image
depth estimation in the wild.

Our contributions are two fold: (1) we propose a new method to
automatically collect high-quality training data for single-view
depth by integrating SfM and a quality assessment network; (2) using this
method we construct YouTube3D, a large-scale dataset
that advances the state of the art of single-view depth estimation
in the wild.

\section{Related Work}

\noindent\textbf{RGB-D from depth sensors}
A large amount of RGB-D data from depth sensors has played a
key role in driving recent research on single-image depth
estimation~\cite{geiger2013vision,silberman2012indoor,chang2017matterport3d,dai2017scannet,schops2017multi}. 
But due to the limitations of depth sensors and the manual effort involved in
data collection, these datasets lack the
diversity needed for arbitrary real world scenes. For example,
KITTI~\cite{geiger2013vision} consists mainly of road scenes; NYU
Depth~\cite{silberman2012indoor}, ScanNet~\cite{dai2017scannet} and
Matterport3D~\cite{chang2017matterport3d}  consist of only indoor
scenes. Our work seeks to address this drawback by focusing on diverse
images in the wild. 

\smallskip\noindent\textbf{RGB-D from computer graphics}
RGB-D from computer graphics is an attractive option because the depth
will be of high quality and it is easy to generate a large amount. 
Indeed, synthetic data
has been used in computer vision with much success~\cite{groueix2018atlasnet,tulsiani2017factoring,mayer2016large,tatarchenko2016multi,Butler:ECCV:2012,fan2017point,choy20163d,wu2016learning,rematas2018soccer}. 
In particular, SUNCG~\cite{song2016semantic} has been shown to improve 
single-view surface normal estimation
on natural indoor images from the NYU Depth dataset~\cite{zhang2017physically}.
However, the diversity of synthetic data is limited by 
the availability of 3D ``assets'', i.e.\@ shapes, materials, layouts, etc., and
it remains difficult to automatically compose diverse scenes
representative of the real world.

\smallskip\noindent\textbf{RGB-D from crowdsourcing}
Crowdsourcing depth annotations~\cite{chen2016single,chen2017surface} has recently received increasing
attention. It's appealing because it can be applied to a truly diverse
set of in-the-wild images. Chen et al.~\cite{chen2016single} crowdsourced annotations of relative depth and
constructed Depth in the Wild (DIW), a large-scale dataset for single-view depth in
the wild. The main drawback of crowdsourcing is, obviously, the cost of manual
labor, and our work attempts to mitigate or avoid this cost through an
automatic method.

\smallskip\noindent\textbf{RGB-D from multiview geometry}
When multiple images of the same scene are available, depth can be
reconstructed through multiview geometry. Prior work has
exploited this fact to collect RGB-D data. 
Xian et al.~\cite{xian2018monocular} perform stereopsis on stereo images, i.e.\@ pairs
 of images taken by two calibrated cameras, to collect a dataset
 called ``ReDWeb''. 
 Li et al.~\cite{li2018megadepth} perform SfM on unordered collections
 of online images of the same scenes to collect a dataset called
 ``MegaDepth''.

Our work differs from prior work in two ways. First, we use a
new source of RGB data---monocular
videos---which likely offer better availability and diversity---stereo
images have limited availability because they must be taken by stereo
cameras. Multiple images of the same scene tend to be biased toward
well-known sites frequented by tourists.

Second, our method of quality assessment is new. Both prior works
performed some form of quality assessment, but neither used
learning. Xian et al.~\cite{xian2018monocular} manually remove
some poor reconstructions; Li et al.~\cite{li2018megadepth} use
handcrafted criteria based on semantic segmentation. In contrast, 
our quality assessment network can learn criteria and patterns beyond
those that are easy to handcraft.

\smallskip\noindent\textbf{Predicting failure} Our work is also
related to prior work on predicting failures for vision
systems~\cite{zhang2014predicting,daftry2016introspective,bendale2016towards,bansal2014towards}.
For example, Zhang et al.~\cite{zhang2014predicting}
predict failure for a variety of vision tasks based solely on the
input. Daftry et al.~\cite{daftry2016introspective} predict failures in 
an autonomous navigation system  directly
from the input video stream. Our method is different in that we predict 
failure in a SfM system to filter reconstructions, based not on the input
images but on the outputs of the SfM system.

\section{Approach}
\label{section:QAnet}

Our method consists of two main steps: SfM followed 
by quality assessment, as illustrated by Fig.~\ref{fig:teaser}. SfM
produces candidate 3D reconstructions, which are then filtered by a
QANet before we use them to generate single-view training data.

\subsection{Structure from Motion}
\label{Section:SfM}
The SfM component of our method is standard. We first detect and
match features across frames. We then estimate the fundamental matrix
and perform triangulation to produce 3D points.

It is worth noting that SfM produces only a sparse reconstruction.
Although we can generate a dense point cloud by a subsequent step of multiview
stereopsis, we choose to forgo it, because
stereopsis in unconstrained settings tends to contain a large amount
of error, especially
in the presence of low-texture surfaces or moving objects.

Our SfM component also involves a couple minor modifications 
compared to a standard full-fledged SfM system. First, we only perform two-view
reconstruction. This is to 
simplify the task of quality assessment---the quality assessment
network only needs to examine two input images as opposed to
many. Second, we do not perform bundle adjustment~\cite{hartley2003multiple}, because we observe
that with unknown focal length of Internet videos (we assume a
centered principal point and focal length is the only unknown
intrinsic parameter), it often leads to poor
results. This is because bundle adjustment is sensitive to
initialization, and tends to converge to an
incorrect local minimum if the initialization of focal length is
not already close to correct. Instead, we search a range of focal
lengths and pick the one that leads to the smallest reprojection error after triangulation. This
approach does not get stuck in local minima, and is justified by the fact that focal length can be uniquely
determined when it is the only unknown intrinsic parameter of a fixed
camera across two views~\cite{pollefeys1999self}.

\subsection{Quality Assessment Network (QANet)}
\label{Section:QANet}
The task of a quality assessment network is to identify good
SfM reconstructions and filter out bad ones. In this section we
discuss important design decisions including the input, output, architecture, and
training of a QANet. 

\paragraph{Input to QANet} The input to a QANet should include 
a variety of cues from the operation of a SfM pipeline on a
particular input. Recall that we consider only two-view
reconstruction; thus the input to SfM is only two video frames.

We consider
cues associated with the entire reconstruction
(reconstruction-wise cues) as well as those
associated with each reconstructed 3D point (point-wise cues). Our
reconstruction-wise cues include the inferred focal length and the average reprojection
error. Our point-wise cues include the 2D coordinates of a feature
match, the Sampson distance of a feature match under the recovered
fundamental matrix, and the angle between the two rays connecting
the reconstructed 3D point and the camera centers.

Note that we do not use any information from the pixel values. The
QANet only has access to geometrical information of the matched
features. This is to allow better generalization by preventing
overfitting to image content. 

Also note that in a SfM pipeline RANSAC is typically used to handle outliers. That
is, multiple reconstructions are attempted on random subsets of the
feature matches. Here we apply the QANet only to the best subset free
from outliers.

\paragraph{Output of QANet} The output of a QANet is a quality score for the
entire reconstruction, i.e.\@ a sparse point cloud. Ideally,
this score should correspond to a similarity metric
between two point clouds,  the reconstructed one and the ground
truth.

There are many possible choices of the similarity metric, with
different levels of invariance and robustness (e.g.\@ invariance to
scale, and robustness to deformation and outliers). Which one
to use should be application dependent and is not the main concern of
this work. And it is sufficient to note that our method is general and not tied to a
particular similarity metric. 

\paragraph{QANet architecture}

\begin{figure}[t]
    \begin{center}
        \includegraphics[width=\linewidth]{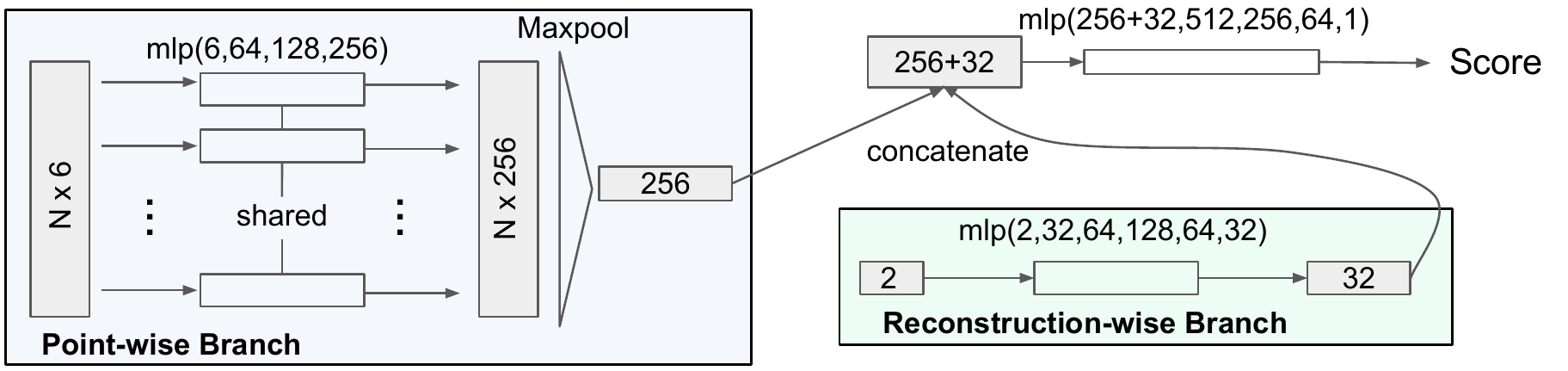}
    \end{center}
    \caption{Architecture of the Quality Assessment Network (QANet).}
    \label{fig:network_architecture}
\end{figure}

Fig.~\ref{fig:network_architecture} illustrates the architecture of our
QANet. It consists of two branches.  The
 \emph{reconstruction-wise branch} processes the reconstruction-wise
 cues (the focal length and overall reprojection error). The \emph{point-wise
   branch} processes features associated with each reconstructed
 point. The outputs from the two branches are then concatenated and fed into
 multiple fully connected layers to produce a quality score. 

Point-wise cues need a separate branch 
because they involve an unordered set of feature vectors with a variable
size. To be invariant to the number and ordering of the vectors, we employ an architecture
 similar to that of PointNet~\cite{qi2016pointnet}. In this
 architecture, each vector is independently processed by shared
 subnetwork and the results are max-pooled at the end.

 \paragraph{QANet training}

 To train a QANet, a straightforward approach is to use a regression
 loss that minimizes the difference between the predicted quality
 score and the ground truth score---the similarity between the
 reconstructed 3D point cloud and the ground truth.

 However, using a regression loss makes
 learning harder than necessary. In fact, 
the absolute value of the score matters much less than the ordering
of the score, because when we use a QANet for filtering, we
remove all reconstructions with scores below a threshold, which can be
chosen by cross-validation. In other words, the 
network just needs to tell that one construction is better than another, but does not need
to quantify the exact degree. Moreover,  the
precision of top-ranked reconstructions is much more important than
the rest, and should be given more emphasis in the loss. 

This observation motivates us to use a ranking loss. Let 
$s_1$ be the ``ground truth quality score'' (i.e.\@ similarity to the
ground truth reconstruction) of a reconstruction in the training set.  Let $s'_1$ be its predicted quality score by the
QANet. Similarly, let $s_2$ be the ground truth quality of another
reconstruction, and let $s'_2$ be the predicted quality score. We
define a ranking loss  $h(s'_1, s'_2, s_1, s_2)$   on this pair of
reconstructions: 
\begin{align}
h(s'_1, s'_2, s_1, s_2) = \left\{
\begin{array}{ll}
\ln\left(1+\exp(s'_2 - s'_1) \right), & \text{if  } s_1 > s_2 \\
\ln\left(1+\exp(s'_1 - s'_2) \right),  & \text{if  } s_1 < s_2 \\
\end{array}
\right.
\label{equation:relative_score}
\end{align}
This loss imposes a penalty if the score ordering of the pair is
incorrect. When applied to all possible pairs, it generates a very large
total penalty if a bad reconstruction is ranked top, because many pairs will
have the wrong ordering. Obviously, in practice we cannot afford to
train with all possible pairs. Instead, we uniformly sample random pairs whose
difference in 
ground truth quality scores are larger than some threshold.

\section{Experiments}

\paragraph{Relative depth}
One implementation question we have left open in the previous sections is the choice
of the ``ground truth'' quality score for the QANet. Specifically, to
train an actual QANet, we
need a similarity metric that compares a reconstructed point cloud
with the ground truth point cloud (the clouds have the same number of
points and known correspondence).

In our experiments we define the similarity metric based on relative
depth. We consider all pairs of points in the reconstructed cloud, and
calculate the percentage of pairs that have the same depth ordering
as the ground truth. Note that depth ordering is view
dependent, and because our SfM component performs two-view reconstruction, we
take the average from both views.

Our choice of relative depth as the quality measure is motivated by two
reasons. First, relative depth is more robust to outliers. Unlike
metrics based on metric difference such as RMSE, with relative depth
a single outlier point will not be able to dominate the error. Second, relative
depth has been used as a standard evaluation metric 
for depth prediction in the
wild~\cite{chen2016single,li2015depth,xian2018monocular,xu2018rendering}, 
partly because it would be difficult to obtain ground truth for
arbitrary Internet images except to use humans, which are good at
annotating relative depth but not metric depth.

Another implementation question is how to train a single-view
depth network with the single-view data generated by our method,
i.e.\@ 3D points from SfM filtered by the QANet. Here we opt to also
derive relative depth from the 3D points. In other words, the final form of our automatically
collected training data is a set
of video frames, each associated with a set of 2D points with their
``ground truth'' depth ordering. 

One advantage of using relative depth as training data is that it is
scale-invariant and 
sidesteps the issue of scale ambiguity in our SfM reconstructions. 
In addition, prior work~\cite{chen2016single} has shown that
relative depth can serve as a good source of supervision even when the
goal is to predict dense metric depth. Last but not least, using
relative depth allows us to compare our automatically
collected data with prior work such as MegaDepth~\cite{li2018megadepth}, which also generates
training data in the form of relative depth.

\subsection{Evaluating QANet}
\label{sec:experiment_QANet}
We first evaluate whether the QANet, as a standalone component, can
be successfully trained to identify high-quality reconstructions. 

We
train the QANet using a combination of existing RGB-D video datasets: NYU
Depth~\cite{silberman2012indoor}, FlyingThings3D~\cite{mayer2016large}, and SceneNet~\cite{mccormac2016scenenet}. We use
the RGB videos to produce SfM reconstructions and use 
the depth maps to compute the ground truth quality score for each
reconstruction.

\begin{figure}[t]
	\begin{center}
		\includegraphics[width=0.9\linewidth]{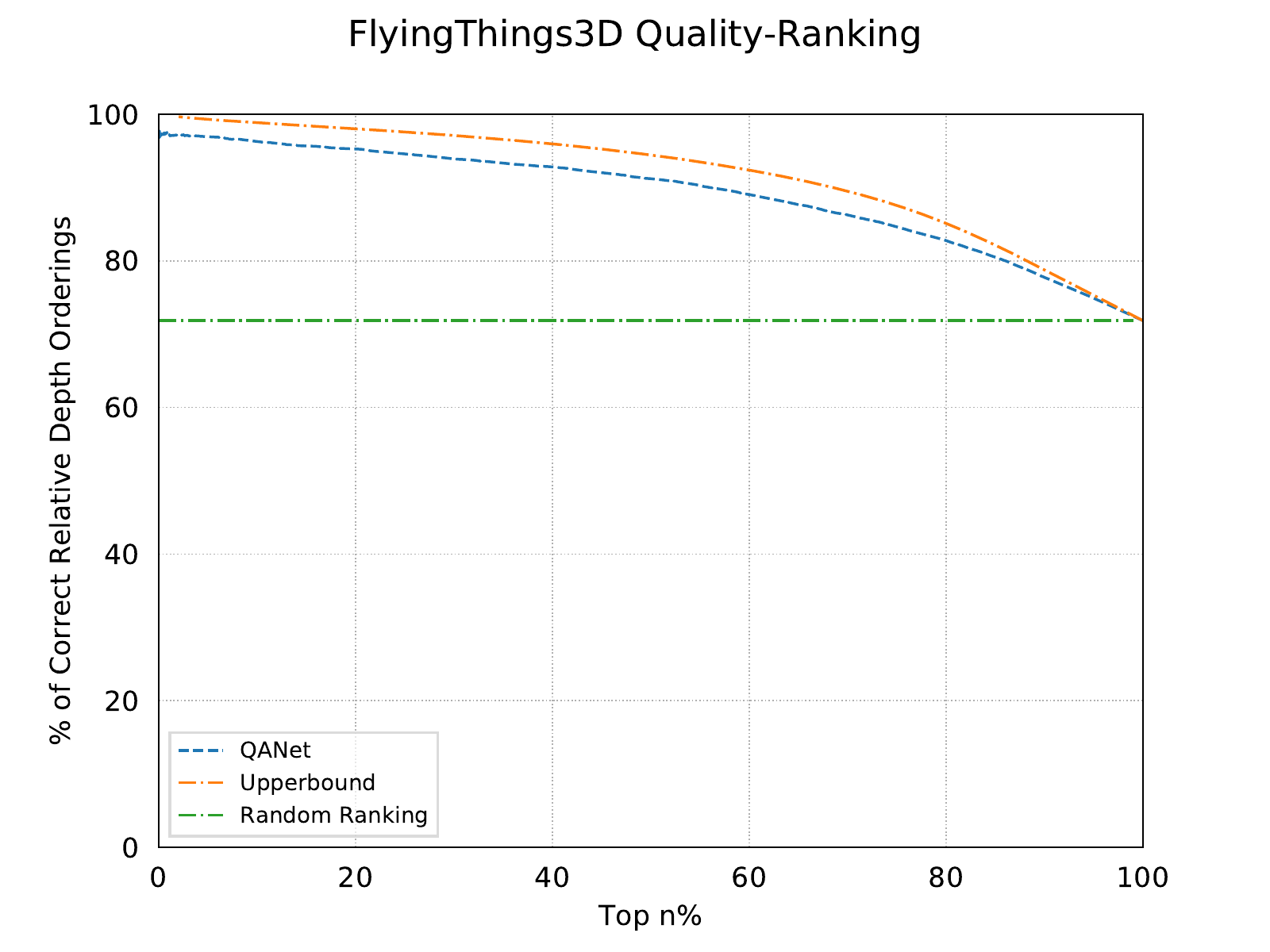}
	\end{center}
	\vspace{-15pt}
	\caption{The quality-ranking curve on the FlyingThings3D dataset.}
	\label{fig:reconstruction_precision_recall_FT3}
\end{figure}

\begin{figure}[t]
    \begin{center}
        \includegraphics[width=0.9\linewidth]{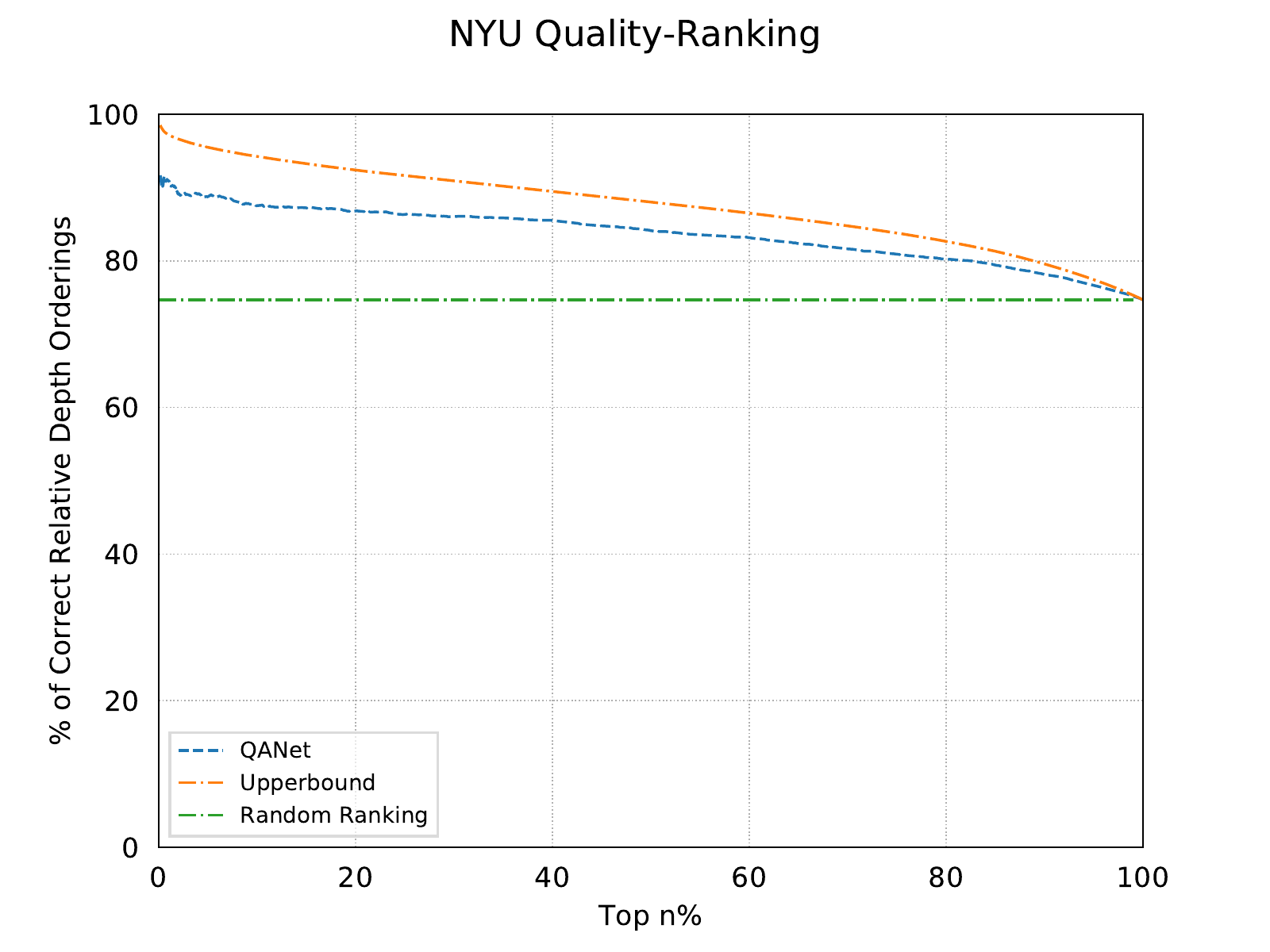}
    \end{center}
    \vspace{-15pt}
    \caption{The quality-ranking curve on the NYU dataset.}
    \label{fig:reconstruction_precision_recall_NYU}
\end{figure}

We measure the performance of our QANet by plotting a quality-ranking
curve---the Y-axis is the average ground-truth quality (i.e.\@
percentage of correct relative depth orderings) of the top $n\%$ reconstructions ranked by QANet, and the X-axis is
the number $n$. At the same $n$, a better QANet would have a better
average quality.

\begin{table}
	\begin{center}
		\begin{tabular}{|c|c|c|}
			\hline
			QANet Variants & \multicolumn{2}{|c|}{AUC} \\ \cline{2-3}			
				&                                 NYU & FlyingThings3D\\ 
			\hline\hline		
			-2D &                             80.53\% & 85.34\%\\
			-Sam&                             83.20\% & 88.66\%\\	
			-Ang&							  82.09\% & 85.00\%\\		
			-Focal&                           82.54\% & 88.37\%\\
			-RepErr      &                    83.37\% & 88.50\%\\
		    \textbf{Full}&           \textbf{83.56\%} & \textbf{89.02\%}\\
			\hline
			Upperbound&                       87.49\% & 91.28\%\\
			Random Ranking&					  75.09\% & 71.41\%\\
			\hline
		\end{tabular}
	\end{center}
	\caption{AUC (area under curve) for different ablated versions of the QANet.}
	\label{table:QANet_Ablative}
\end{table}

We test our QANet on the test splits of FlyingThings3D and NYU
Depth. The results are shown in Fig.~\ref{fig:reconstruction_precision_recall_FT3} and
Fig.~\ref{fig:reconstruction_precision_recall_NYU}. In both figures,
we provide an \emph{Upperbound} curve from a perfect ranking of
the reconstructions, and a \emph{Random Ranking} curve from a random
ranking of the reconstructions.

From Fig.~\ref{fig:reconstruction_precision_recall_FT3} and
Fig.~\ref{fig:reconstruction_precision_recall_NYU} we see that our QANet
can successfully rank reconstructions by quality. On FlyingThings3D, 
the average quality of unfiltered (or randomly ranked) reconstructions
is 71.41\%, whereas the top 20\% reconstructions ranked by QANet have an average
quality of 95.26\%. On
NYU Depth, the numbers are 75.09\% versus 86.80\%.

In addition, we see that the QANet curve is quite close to the
upperbound curve. On FlyingThings3D, the AUC (area under curve) of the
upperbound curve is 91.28\%, and the AUC of QANet is 89.02\%. On NYU
Depth, the numbers are 87.49\% and 83.56\%.

\paragraph{Ablative Studies}
We next study the contributions of different cues to quality
assessment. We train five ablated versions of QANet by 
(1) removing 2D coordinate feature (-2D);
(2) removing Sampson distance feature (-Sam);
(3) removing angle feature (-Ang); 
(4) removing focal length (-Focal);
(5) removing reprojection error (-RepErr).

We compare their performances in terms of AUC with the full QANet in Tab.~\ref{table:QANet_Ablative}.
They all underperform the full QANet, indicating that all cues contribute
to successful quality assessment.

\begin{figure*}[t]
	\begin{center}
		\includegraphics[width=\linewidth]{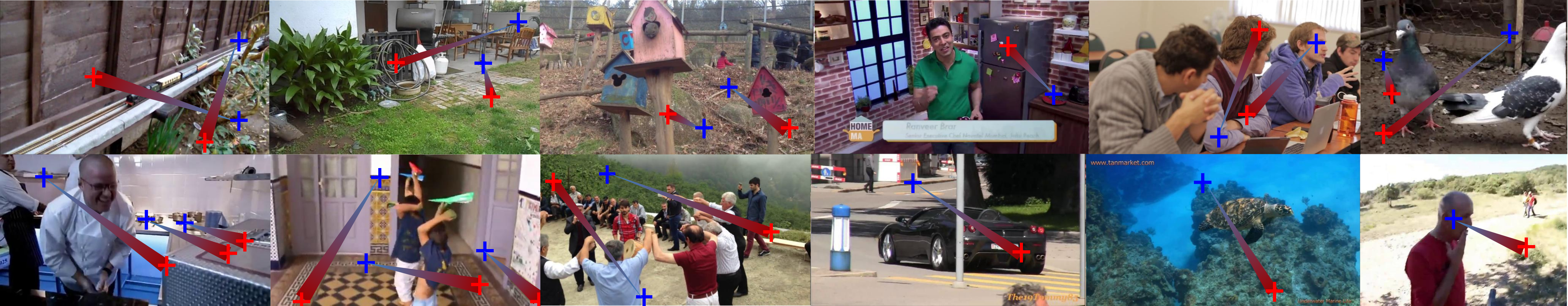}
	\end{center}
	
	\caption{Examples of automatically collected relative depth
          annotations in YouTube3D. The
          relative depth pairs are visualized as two connected points,
          with red point being closer than the blue point. These
          relative depth annotations are mostly correct. 
	}
	\label{fig:demo_YT3D}
\end{figure*}

\subsection{Evaluating the full method}

We now turn to evaluating our full data collection method. To this
end, we need a way to compare our dataset with those collected by alternative methods. 

Note that it is insufficient to compare datasets using the accuracy
of the ground truth labels, because the datasets may have
different numbers of images, different images, or different annotations
on the  same images (e.g.\@ different pairs of points for relative depth). A
dataset may have less accurate labels, but may still end up
more useful due to other reasons such as better diversity or more
informative annotations.

Instead, we compare datasets by their usefulness for training. In
our case, a dataset is better if it trains a better deep network for
single-view depth estimation. Given a dataset of relative depth, we
use the method of Chen et al.~\cite{chen2016single} to train 
a image-to-depth network by imposing a ranking loss on the output depth values
to encourage agreement with the ground truth orderings. We measure the
performance of the trained network by the weighted human disagreement
rate (\textit{WHDR})~\cite{chen2016single}, i.e.\@ the percentage of
incorrectly ordered point pairs.

\vspace{-4mm}
\paragraph{YouTube3D}

We crawled 0.9 million YouTube videos using random keywords.
Pairs of frames	are randomly sampled and selected if feature matches exist
between them. We apply our method to these pairs and obtain
2 million filtered reconstructions spanning 121,054 videos.  From
 these reconstructions we construct a dataset called \emph{YouTube3D},
 which consists of 795,066 images, with an average of 281 relative
 depth pairs per image. Example images
 and annotations of YouTube3D are shown in Fig.~\ref{fig:demo_YT3D}. 
 
As a baseline, we construct another dataset called \emph{YT$_{UF}$}.
It is built from all reconstructions that are used in constructing YouTube3D 
but without applying the QANet filtering. Note that YT$_{UF}$ is a 
superset of YouTube3D, and contains 3.5M images. 

\vspace{-3mm}
\paragraph{Colmap}
Our implementation of SfM is adapted from Colmap~\cite{schoenberger2016sfm}, a
state-of-the-art SfM system. We use the same feature matches generated by
Colmap, and modified the remaining steps as described in Sec.~\ref{Section:SfM}.  
In our experiments, we also include the
original unmodified Colmap system as a baseline. To generate relative depth from the sparse point
clouds given by Colmap, we randomly sample point pairs
and project them into different views.

We run Colmap on the same set of features and matches as used in constructing 
YouTube3D and YT$_{UF}$, obtaining 647,143 reconstructions that span 486,768 videos.
From them we construct a dataset called \emph{YT$_{Col}$}. It contains 3M images,
with an average of 4,755 relative depth pairs per image. 

\vspace{-3mm}
\paragraph{Depth-in-the-Wild (DIW)} 
We use the Depth-in-the-Wild (DIW) dataset~\cite{chen2016single} to evaluate the
performance of a single-view depth network. DIW consists of Internet
images that cover diverse types of scenes. It has 74,000
test and 420,000 train images; each image has human
annotated relative depth for one pair of points. 
In addition to using
the test split of DIW for evaluation, we also use its training split as a standalone
training set.

\begin{table}
	\begin{center}
			\begin{tabular}{|c|c|}
				\hline
				Training Sets &  WHDR \\
				\hline\hline			
				NYU	&31.31\%~\cite{chen2016single}\\
				DIW	&22.14\%~\cite{chen2016single} \\		
	            MegaDepth&  22.97\%~\cite{li2018megadepth}\\	
				YT$_{Col}$ & 34.47\%\\	
				YT$_{UF}$  & 25.11\%\\	
				QA\_train  & 31.77\% \\    
				NYU + QA\_train & 31.22\% \\			
				\textbf{YouTube3D}  & \textbf{19.01\%}\\	
				\hline
			\end{tabular}
	\end{center}	
	\caption{Error rate on the DIW test set by the Hourglass Network~\cite{chen2016single} trained on different standalone datasets.}
	\label{table:comparison_as_standalone_HG}	
\end{table}

\begin{figure*}[ht]
	\begin{center}
		\includegraphics[width=\linewidth]{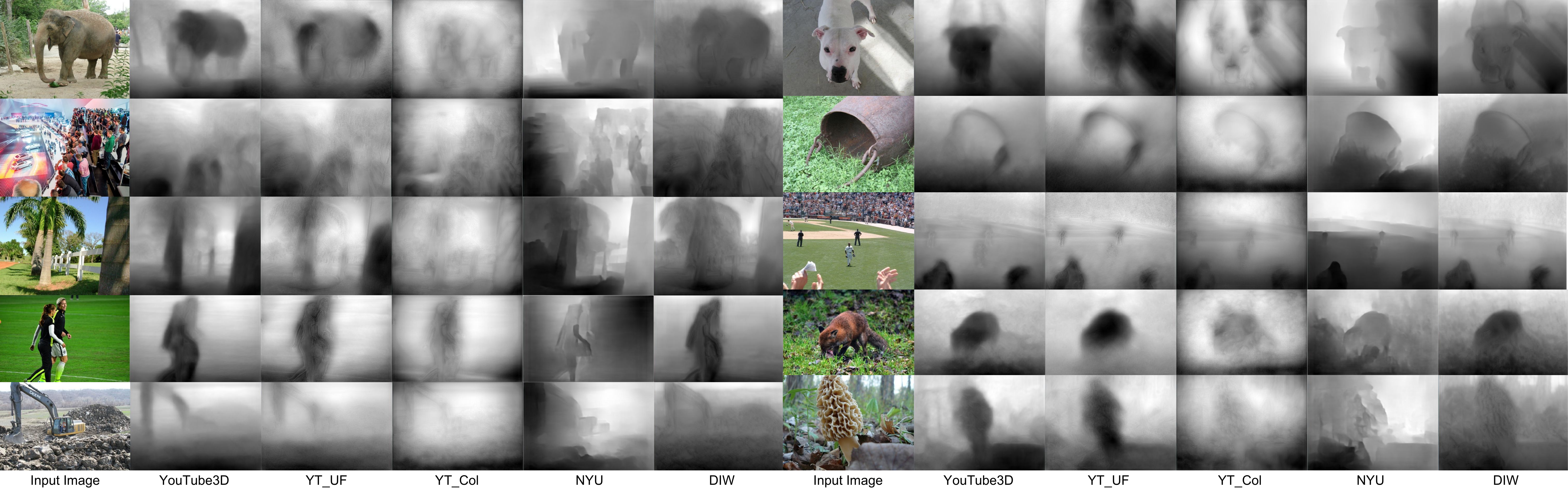}
	\end{center}
	\vspace{-4mm}
	\caption{Qualitative results on the DIW test set by the Hourglass Network~\cite{chen2016single} trained with different datasets. Column names denote the datasets used for training. }
	\label{fig:qual_result_HG}
\end{figure*}

\vspace{-3mm}
\paragraph{Evaluation as standalone dataset}

We evaluate YouTube3D as a standalone dataset and compare it
with other datasets. That is, we train a single-view depth network from
scratch using each dataset and measure the performance on DIW.
To directly compare with existing results in the literature, 
we use the same hourglass network that has been used in a number of prior
works~\cite{chen2016single,li2018megadepth}. 

Tab.~\ref{table:comparison_as_standalone_HG} compares the DIW performance of a hourglass network trained on YouTube3D 
against those trained on three other datasets:  MegaDepth~\cite{li2015depth}, NYU
Depth~\cite{silberman2012indoor}, and the training split of DIW~\cite{chen2016single}.
The results are shown in Tab.~\ref{table:comparison_as_standalone_HG}.
We see that YouTube3D not only outperforms NYU Depth, 
which was acquired with depth sensors, but also MegaDepth, another high-quality
depth dataset collected via SfM. 
Most notably, even though the evaluation is on DIW, YouTube3D outperforms the 
training split of DIW, showing that our automatic data collection
method is a viable substitute for manual annotation.

Tab.~\ref{table:comparison_as_standalone_HG} also compares YouTube3D
against YT$_{UF}$ (YouTube3D without QANet filtering) and
YT$_{Col}$ (off-the-shelf SfM). We see that YouTube3D outperforms 
the unfiltered set YT$_{UF}$ by a large margin, even though
YT$_{UF}$ is a much larger superset of YouTube3D. This underscores
the effectiveness of QANet filtering. Moreover, YouTube3D outperforms 
YT$_{Col}$ by an even larger margin, indicating our method is much
better than a direct application of off-the-shelf state-of-the-art SfM
to Internet videos. Notably, YT$_{UF}$ already outperforms YT$_{Col}$ significantly. This is a 
result of our modifications described in 
Sec.~\ref{Section:SfM}: (1) we require the estimate of the fundamental matrix
to have zero outliers during RANSAC; (2) we replace bundle adjustment
with a grid-search of focal length.

Fig.~\ref{fig:qual_result_HG} shows a qualitative comparison of depth estimation by
networks trained with different datasets. We can see that training on
YouTube3D generally produces better results than others, especially compared to
$YT_{Col}$ and NYU.

We also include a comparison between YouTube3D and \emph{QA\_train}, the
data used to train QANet. This is to answer the question whether a
naive use of this extra data---using it directly to train a
single-view depth network---would give the same advantage enjoyed by
YouTube3D, rendering our method unnecessary. 
We see in Tab.~\ref{table:comparison_as_standalone_HG} that
training single-view depth directly from  QA\_train is much worse than YouTube3D (31.77\%
vs. 19.01\%), showing that QA\_train itself is a
not a good training set for mapping pixels to depth. In addition, adding
QA\_train to NYU Depth (NYU + QA\_train in Tab.~\ref{table:comparison_as_standalone_HG})
barely improves the performance of NYU Depth alone. This shows that a
naive use of this extra data will not result in the improvement
achievable by our method. It also shows that QANet generalizes well to
images in the wild, even when trained on data that is quite different
in terms of pixel content. It is worth noting that this result should
not be surprising, because QANet does not use pixel values to assess
quality and only uses the geometry of the feature matches.

\begin{table}
	\begin{center}
		\resizebox{\columnwidth}{!}{
			\begin{tabular}{|c|c|c|}
				\hline
				Network & Training Sets &  WHDR \\
				\hline\hline	
				Hourglass&NYU + DIW	 & 	14.39\%~\cite{chen2016single}\\
				\cite{chen2016single}&\textbf{NYU + DIW + YouTube3D}& \textbf{13.50\%} \\				
				\hline	
EncDecResNet			&ImageNet + ReDWeb& 14.33\%\\
\cite{xian2018monocular}&ImageNet + ReDWeb + DIW & 11.37\%\\				
				\hline
EncDecResNet			&ImageNet + ReDWeb& 16.31\%\\
(Our Impl  &ImageNet + YouTube3D & 16.21\%\\ \cline{2-3}
of~\cite{xian2018monocular}) &ImageNet + ReDWeb + DIW & 12.03\%\\				
&\textbf{ImageNet + ReDWeb + DIW + YouTube3D} & \textbf{10.59\%}\\
				\hline
			\end{tabular}
		}
	\end{center}	
	\caption{Error rate on the DIW test set by networks trained with and without YouTube3D as supplement.}
	\label{table:comparison_as_suppl}	
\end{table}

\vspace{-4mm}
\paragraph{Evaluation as supplemental dataset}
We evaluate YouTube3D as supplemental data. Prior works have
demonstrated state-of-the-art performance on DIW by combining multiple sources of
training data~\cite{chen2016single,xian2018monocular}. We investigate whether
adding YouTube3D as additional data would improve state-of-the-art
systems.

We first add YouTube3D to NYU + DIW, the combined training set used by 
Chen et al.~\cite{chen2016single} to train the first
state-of-art system for single-view depth in the wild. We train the
same hourglass network used in ~\cite{chen2016single}. 
Results in Tab.~\ref{table:comparison_as_suppl} show that with the
addition of YouTube3D, the network is able to achieve a significant
improvement.

\begin{figure*}[ht]
	\begin{center}
		\includegraphics[width=\linewidth]{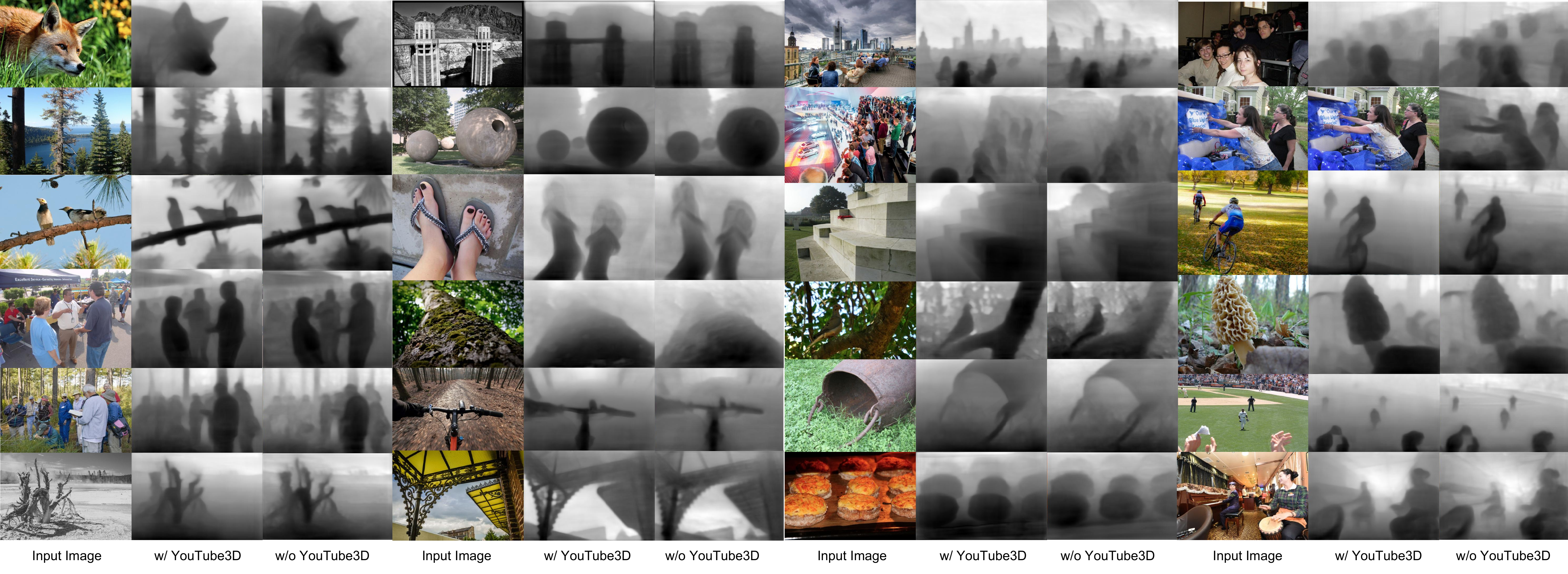}
	\end{center}
	\vspace{-4mm}
	\caption{Qualitative results on the DIW test set by the EncDecResNet~\cite{chen2016single} trained on ImageNet + ReDWeb + DIW (\textit{w/o YouTube3D}), and fine-tuned  on YouTube3D (\textit{w/ YouTube3D}). }
	\label{fig:qual_result_ResNet}
\end{figure*}

We next evaluate whether YouTube3D can improve the best existing result on
DIW, achieved by an encoder-decoder network based on
ResNet50~\cite{xian2018monocular} (which we will refer to as
an EncDecResNet subsequently). The network is trained on a
combination of ImageNet, DIW, and ReDWeb, a relative depth dataset
collected by performing stereopsis on stereo images with manual removal of
poor-quality reconstructions. Tab.~\ref{table:comparison_as_suppl} summarizes our
results, which we elaborate below. 

We implement our own version of the EncDecResNet used in
~\cite{xian2018monocular}, because there is no public code available
as of writing. As a validation of our implementation, we train the
network on ImageNet and ReDWeb, and achieve an error rate
of $16.31\%$, which is slightly worse than but sufficiently close to the $14.33\%$ 
reported in ~\cite{xian2018monocular}\footnote{All results in
  ~\cite{xian2018monocular} are with ImageNet.}.  This discrepancy is likely
because certain details (e.g.\@ the exact number of channels at each
layer) are different in our implementation because they are not
available in their paper.

As an aside, we train the same EncDecResNet on ImageNet and YouTube3D,
which gives an error rate of $16.21\%$, which is comparable with the
$16.31\%$ given by ImageNet and ReDWeb. This suggests that YouTube3D
is as useful as ReDWeb. This is noteworthy because unlike ReDWeb,
YouTube3D is not restricted to stereo images and does not involve any manual
filtering. Note that it is not meaningful to compare with the
$14.33\%$ reported in ~\cite{xian2018monocular}---to compare two
training datasets we need to train the exact same network, but the
$14.33\%$ is likely from a slightly different network due to the
unavailability of some details in ~\cite{xian2018monocular}.

Finally, we train an EncDecResNet on the
 combination of ImageNet, DIW, and ReDWeb, which has produced the current state of the art
on DIW in ~\cite{xian2018monocular}. With our own implementation we achieve an error rate of $12.03\%$, slightly worse than
the $11.37\%$ reported in ~\cite{xian2018monocular}. Adding YouTube3D to the
mix, we achieve an error rate of $10.59\%$, a new state of the art
performance on DIW (see Fig.~\ref{fig:qual_result_ResNet} for example depth estimates). This result demonstrates the effectiveness of
YouTube3D as supplemental single-view training data. 

\vspace{-4mm}
\paragraph{Discussion}

The above results suggest that our proposed method can generate high-quality
training data for single-view depth in the wild. Such results are
significant, because our dataset is gathered by a \emph{completely automatic}
method, while datasets like DIW~\cite{chen2016single} and ReDWeb
~\cite{xian2018monocular} are constrained by manual
labor and/or the availability of stereo images. Our automatic method
can be readily applied to a much larger set of Internet videos and
thus has potential to advance the state of the art of single-view
depth even more significantly. 

\vspace{-2mm}
\section{Conclusion}
In this paper we propose a fully automatic and scalable method for collecting
training data for single-view depth from Internet videos. Our method performs SfM and uses a Quality
Assessment Network to find high-quality reconstructions, which are used to produce single-view
depth ground truths. We apply the proposed method on YouTube videos and construct
a single-view depth dataset called YouTube3D. We show that YouTube3D is useful both 
as a standalone and as a supplemental dataset in training depth predictors. With it, we
obtain state-of-the-art results on single-view depth estimation in the wild.

\vspace{-2mm}
\section{Acknowledgment}
This publication is based upon work partially supported by the King Abdullah University of Science 
and Technology (KAUST) Office of Sponsored Research (OSR) under Award No. OSR-2015-CRG4-2639, a gift 
from Google, and the National Science Foundation under grant 1617767.

{\small
\bibliographystyle{ieee}
\bibliography{egbib}
}

\end{document}